\documentclass{article}

\usepackage[preprint]{neurips_2025}

\usepackage[utf8]{inputenc} 
\usepackage[T1]{fontenc}    
\usepackage{hyperref}       
\usepackage{url}            
\usepackage{multirow}
\usepackage{booktabs}       
\usepackage{amsfonts}       
\usepackage{nicefrac}       
\usepackage{microtype}      
\usepackage{xcolor}         
\usepackage{makecell}
\usepackage{graphicx}
\usepackage{array}
\usepackage{amsmath}
\usepackage{caption}
\usepackage{mwe}

\title{\name{}: A Foundation Model for Physics Simulations}

\author{%
    Tung Nguyen\thanks{Equal contribution.} \\
    UCLA \\
    \texttt{tungnd@cs.ucla.edu}
    \And
    Arsh Koneru\footnotemark[1] \\
    UCLA \\
    \texttt{arshkon@g.ucla.edu} \\
    \AND
    Shufan Li\footnotemark[1] \\
    UCLA \\
    \texttt{jacklishufan@cs.ucla.edu} 
    \And  
    Aditya Grover \\
    UCLA \\
    \texttt{adityag@cs.ucla.edu}
}

\newlength{\imwid}
\newlength{\imht}
\newcommand{\name}{PhysiX}

\begin{document}

\maketitle
\begin{abstract}
Foundation models have achieved remarkable success across video, image, and language domains. By scaling up the number of parameters and training datasets, these models acquire generalizable world knowledge and often surpass task-specific approaches. However, such progress has yet to extend to the domain of physics simulation. A primary bottleneck is data scarcity: while millions of images, videos, and textual resources are readily available on the internet, the largest physics simulation datasets contain only tens of thousands of samples. This data limitation hinders the use of large models, as overfitting becomes a major concern. As a result, physics applications typically rely on small models, which struggle with long-range prediction due to limited context understanding. Additionally, unlike images, videos, or text—which typically exhibit fixed granularity—physics datasets often vary drastically in scale, amplifying the challenges of scaling up multitask training. We introduce \textbf{\name{}}, the first large-scale foundation model for physics simulation. \name{} is a 4.5B parameter autoregressive generative model. It uses a discrete tokenizer to encode physical processes at different scales into a sequence of discrete tokens, and employs an autoregressive next-token prediction objective to model such processes in the token space. To mitigate the rounding error in the discretization process, \name{} incorporates a specialized refinement module. Through extensive experiments, we show that \name{} effectively addresses the data bottleneck, outperforming task-specific baselines under comparable settings as well as the previous absolute state-of-the-art approaches on The Well benchmark. Our results indicate that knowledge learned from natural videos can be successfully transferred to physics simulation, and that joint training across diverse simulation tasks enables synergistic learning. Code is available at \href{https://github.com/ArshKA/PhysiX}{https://github.com/ArshKA/PhysiX}
\end{abstract}

%



\section{Introduction}
Simulating physical systems governed by partial differential equations (PDEs) is a cornerstone of modern science and engineering. From modeling climate and fluid dynamics to understanding galaxy formation and biological morphogenesis, PDE-based simulations enable us to predict, control, and optimize complex natural phenomena~\citep{eyring2016overview,berger2024implicit,biegler2003large,mohammadi2004shape,cranmer2020frontier,lemos2023simbig}. Traditionally, physics simulations have relied on numerical solvers that discretize and integrate governing equations over space and time. While highly accurate, such methods are computationally intensive, often requiring specialized hardware and expert-tuned software~\citep{goldberg2022numerical}. This high cost has led to growing interest in machine learning (ML)-based surrogates, which aim to approximate simulation outputs at a fraction of the expense~\citep{sun2020surrogate,tao2019application,haghighat2021physics}. Recent work has shown that deep neural networks can learn surrogate models for a range of PDE-driven systems, enabling orders-of-magnitude reductions in inference time~\citep{torlai2018neural,ryczko2019deep,choudhary2022recent,siahkoohi2023martian,gopakumar2024plasma}.

Despite these promising advances, current ML-based surrogates remain largely task-specific. Most methods are designed for a single physical system and trained from scratch using individual datasets. These models typically struggle to adapt when simulation parameters change, such as domain geometry, boundary conditions, or physical constants, and often require significant retraining or architectural modification to maintain accuracy~\citep{franco2023deep,zhang2023label,nguyen2024efficient,gupta2022towards}. Moreover, since they are trained separately for each task, they fail to capture shared inductive biases across domains, such as spatiotemporal locality, symmetry, or conservation laws. 
To address similar limitations of task-specific models in other domains, researchers have increasingly adopted the foundation model paradigm, where a large model is first pretrained on a large set of diverse data, before being finetuned for specific tasks~\citep{brown2020language,bommasani2021opportunities}.


\begin{figure}[t]
    \centering
    \includegraphics[width=0.97\linewidth]{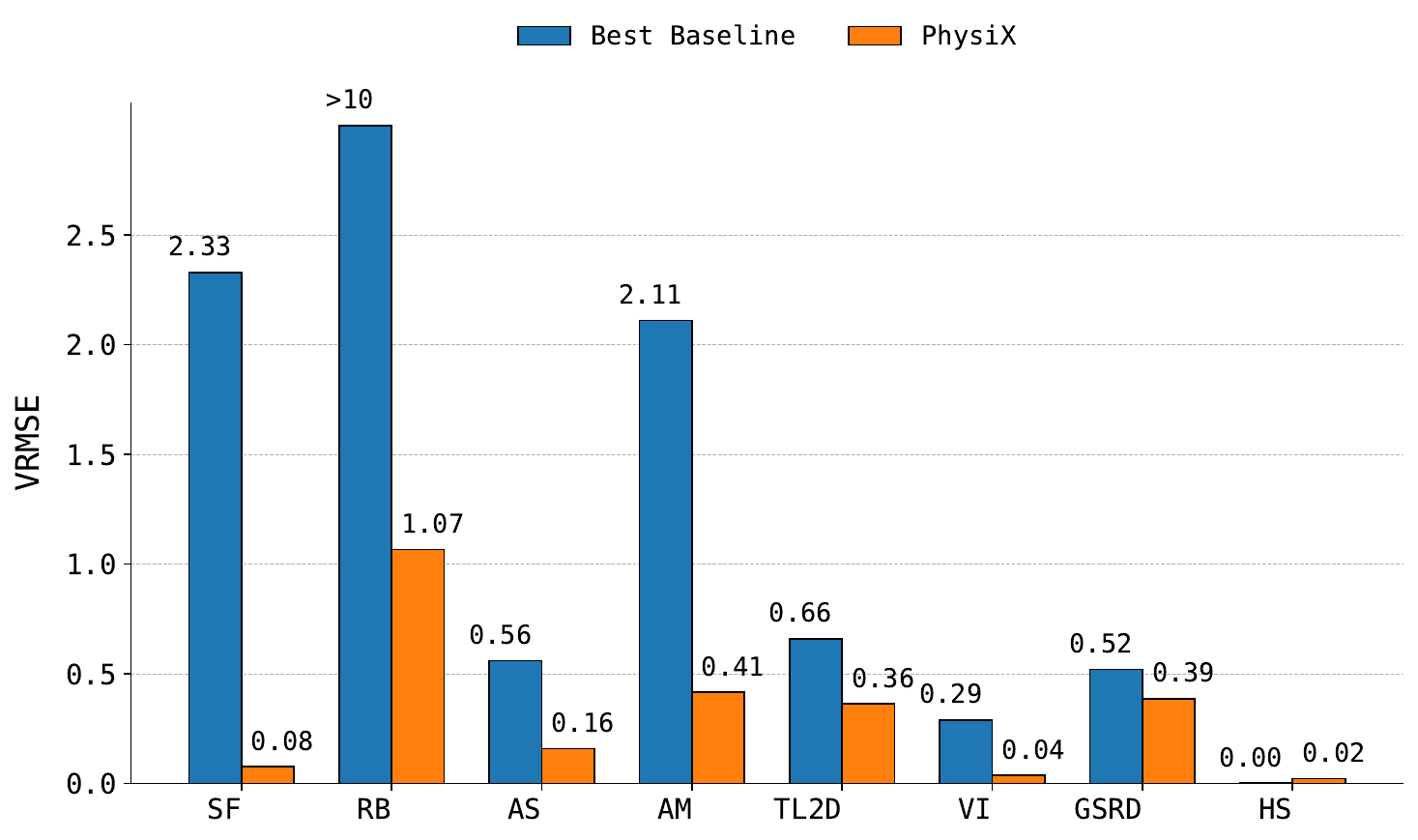}
    \caption{We propose \textbf{\name{}}, a foundation model pretrained for physics simulations. We train \name{} over a collection of 8 physics simulation tasks of the Well benchmark, resulting in a multi-task model that outperforms previous single-task baselines. We report VRMSE (lower is better) averaged across different physical properties and lead time between 9-26 frames for each task.}
    \label{fig:teaser}
\end{figure}

The success of foundation models raises a natural question: \emph{can we build a foundation model for physical simulations?} Unlike text or images, physics simulations pose unique challenges. First, simulation data is expensive to generate and inherently limited in volume. Even the largest public datasets contain only tens of thousands of spatiotemporal examples~\citep{ohana2024well}, orders of magnitude smaller than the text or video corpora used to train large language and vision models. In addition, physical systems exhibit substantial diversity, varying in resolution, dimensionality, underlying equations, and physical domains -- from turbulent fluids to elastic solids and chemical reaction-diffusion systems. Modeling such heterogeneity requires a flexible architecture and a training strategy capable of learning shared representations across domains while preserving task-specific fidelity. Together, these challenges make it non-trivial to scale the foundation model paradigm to physical simulations.

In this work, we introduce \textbf{\name{}}, the first large-scale autoregressive foundation model for physical simulations. \name{} comprises three main components: a universal discrete tokenizer, a 4.5B parameter autoregressive transformer, and a refinement module. We first train the tokenizer jointly on a diverse collection of physics simulation datasets to compress continuous spatiotemporal fields at different scales into sequences of discrete tokens, allowing the model to capture shared structural and dynamical patterns across domains. This discrete representation enables effective data fusion across heterogeneous sources and allows training on a unified token space analogous to language modeling. Building on this representation, we then train a large-scale autoregressive transformer using a next-token prediction objective over the combined tokenized corpus. To further improve generalization and mitigate data scarcity, we initialize both the tokenizer and the autoregressive model from pretrained checkpoints of high-capacity video generation models, enabling \name{} to leverage strong spatiotemporal priors learned from natural videos. Finally, to address the quantization error introduced by tokenization and improve output fidelity, \name{} incorporates a lightweight refinement module that reconstructs fine-scale details from predicted token sequences.

Empirically, we find that \name{}~significantly outperforms task-specific baselines and previous state-of-the-art models on The Well benchmark~\citep{ohana2024well}, demonstrating improved long-range prediction and better generalization across tasks. Figure~\ref{fig:teaser} highlights these results. Our experiments show that \name{} can effectively transfer knowledge from natural video pretraining to physics simulations, and that joint training across multiple simulation datasets enables synergistic learning. These results demonstrate the first compelling evidence that foundation models can serve as unified surrogates for diverse physical systems, bringing us closer to general-purpose, scalable, and efficient tools for scientific computing.

\section{Related Works}

\textbf{Physics Simulation }
Traditional simulation modeling typically relies on numerical methods, such as finite element methods, finite difference methods, and finite volume methods, to approximate solutions to differential equations governing physical laws. While effective, these approaches often require significant computational resources, especially for high-resolution simulations or long-term predictions, limiting their scalability and real-time applicability.

Advances in machine learning have offered promising alternatives to accelerate or supplement traditional PDE solvers \cite{subramanian2023towards, karniadakis2021physics_informed_review}. Physics-informed neural networks (PINNs) incorporate prior knowledge of governing equations into the loss function \cite{raissi2019physics}. These methods require little observational data, as physical constraints guide the learning process. This provides the benefit of interpretable and improved physical plausibility, but makes PINNs an unsuitable choice when the underlying physical laws are unknown or only partially understood.

Concurrently, data-driven surrogate modeling methods have also seen success in this area, shifting from explicitly modeling physical laws towards implicitly learning system dynamics through observed data \cite{lu2019deeponet}. Early work utilized CNNs, particularly U-Net architectures \cite{ronneberger2015u, zhu2018bayesian}, to model spatiotemporal relationships between physical fields. More recently, neural operator frameworks have emerged, which aim to learn mappings between infinite-dimensional function spaces \cite{kovachki2023neural, lu2021learning}. These include Fourier Neural Operators (FNOs) \cite{li2021fourier}, which leverage Fast Fourier Transforms for efficient global convolution, and various Transformer-based architectures \cite{li2022transformer, kissas2022learning} that utilize attention mechanisms to capture long-range dependencies. To handle complex  geometries where methods like FNOs may struggle, Graph Neural Network (GNN) based operators have also been developed, capable of operating directly on unstructured meshes \cite{li2020neural, brandstetter2022message}. These operator learning frameworks enable generalization to different initial conditions, boundary conditions, and spatial resolutions without explicit retraining.

Despite these advancements, current neural network based physics simulators face limitations. They often struggle on long-range predictions \cite{lippe2023pde} and many models are specialized, typically trained and optimized for a specific physical system, a narrow range of parameters, or a particular set of governing equations. Current neural network approaches are able to generalize within a given physical domain, but generally do not perform well across distinct physical domains without substantial retraining or architectural modifications.

\textbf{Video Generation }
Video generation models have achieved considerable progress in recent years. \cite{wang2025wan,kong2024hunyuanvideo,openai2024sora,agarwal2025cosmos}. These models achieve high-fidelity video generation by pre-training on web-scale video data \cite{abu2016youtube,bain2021frozen}. The most common approach for video generation employs diffusion models \cite{ho2022video,bar2024lumiere,wang2025lavie}, which model videos in a continuous latent space. Several works also explored autoregressive video modeling \cite{kondratyuk2023videopoet, gafni2022make}, which convert videos into sequences of discrete tokens using a discrete tokenizer and apply the next-token prediction objective. Most notably, Emu3 \cite{wang2024emu3} demonstrated that autoregressive models can achieve competitive performance when compared with diffusion models at scale. There are several dedicated lines of work focusing on specific design choices of video generative models, including video tokenizer \cite{yu2023language,wang2024omnitokenizer}, model architecture \cite{qing2024hierarchical}, and learning objective \cite{sun2025ar}.

\begin{figure}[t]
    \centering
    \includegraphics[width=1.0\linewidth]{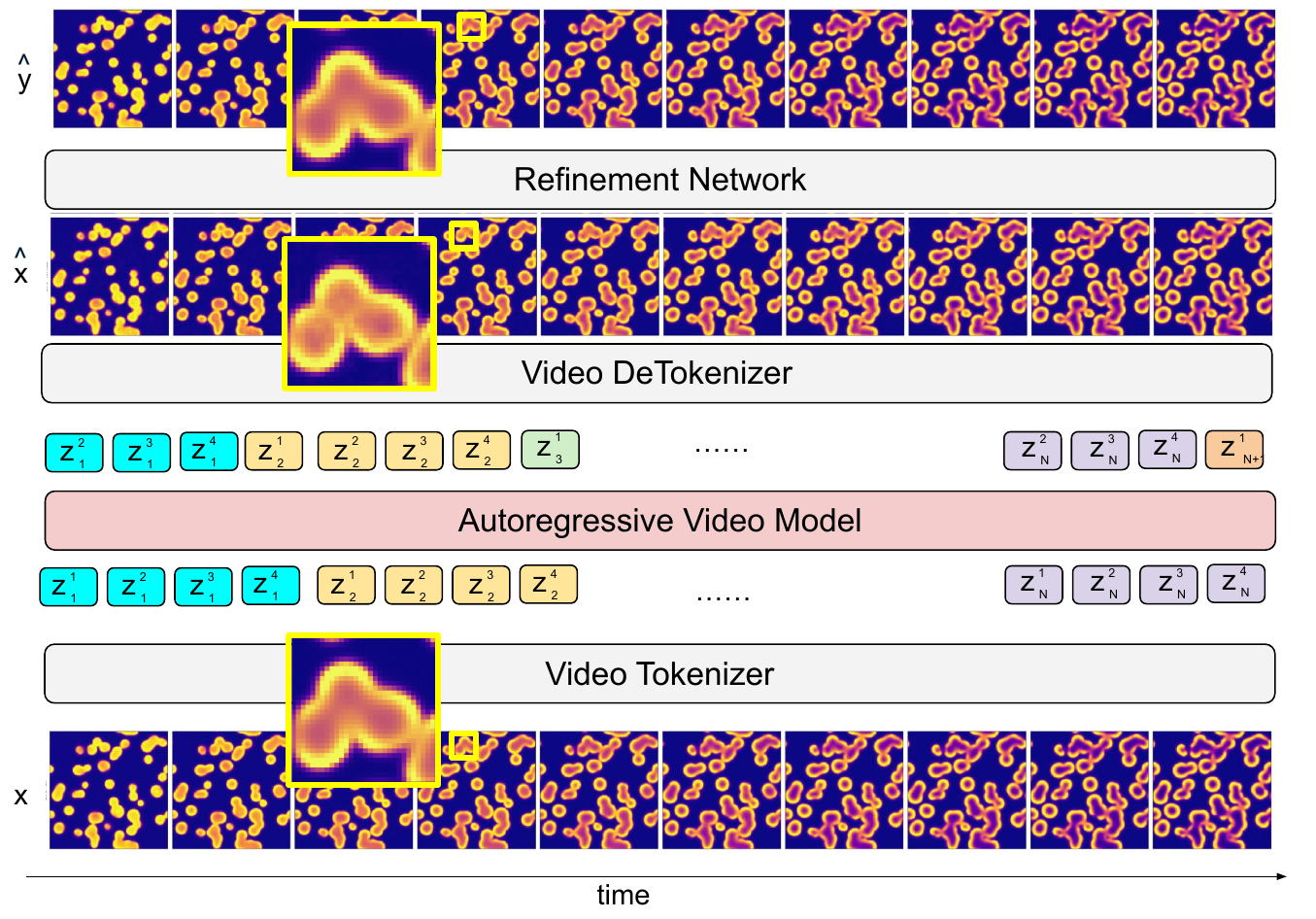}
    \caption{\textbf{The overall design of \name{}}. \name{}~consists of a video tokenizer, an autoregressive model, and a refinement network. Given input frames $x_1,\dots,x_N$, the tokenizer discretizes each frame into a sequence of discrete tokens, where the $j$th token of frame $i$ is denoted as  $\{z_i^j\}$. The autoregressive model then generates predictions in this discrete token space, which are converted back to pixel-level predictions $\hat{x}$ by the de-tokenizer. A refinement module is incorporated to mitigate artifacts caused by the discretization error, such as blocky, pixelated outputs (visualized in yellow boxes), and produce the final sharper and more detailed output $\hat{y}$. }
    \label{fig:explainer}
\end{figure}

\textbf{Foundation Models }
The concept of foundation models first emerged in the context of transfer learning \cite{zhuang2020comprehensive}, where a model trained on large-scale data in one domain can be easily fine-tuned to perform many tasks in adjacent domains. Notable early examples include self-supervised learning on ImageNet-1K, a dataset of natural images \cite{chen2020simple,he2020momentum,oquab2023dinov2}. These pre-trained vision models proved to be versatile for a wide range of downstream applications such as medical imaging \cite{kim2022transfer}. More recent works shifted the training paradigm to vision-language alignment. Models like CLIP \cite{radford2021learning} are pre-trained on large amounts of image-text pairs and have demonstrated strong zero-shot generalization capabilities to a wide range of downstream tasks across multiple domains. Most recently, several works have focused on building foundation models for domain-specific use cases such as remote sensing \cite{reed2023scale}, weather forecasting \cite{nguyen2023climax}, and material design \cite{takeda2023multi}. Most notably, Cosmos \cite{agarwal2025cosmos} builds a foundation world model for physical AI by pre-training on large amounts of video documenting physical applications using the video modeling objective. Its training data covers a wide range of physical applications such as robotic manipulation and self-driving. In this work, we investigate if similar approaches can be adapted to build a foundation model for physics simulations.
\section{Method}

\name{} consists of three components: a discrete tokenizer, an autoregressive generation model, and a refinement module. Given $k$ input frames $x_1,x_2,\dots,x_k$ as the historical context, we first convert them into sequences of $\hat{k}$ latent discrete tokens $z_1,z_2,\dots,z_{\hat{k}}$, where each $z_i=[z_i^1,z_i^2,\dots,z_i^L]$ consists of $L$ tokens. We then concatenate these sequences into a single sequence, and use it as the input for the autoregressive model. The autoregressive model then predicts discrete tokens, $z_{\hat{k}+1},z_{\hat{k}+2},z_{\hat{k}+\hat{T}}$  where $\hat{T}$ is the maximum lead time measured in latent frames (corresponding to $T$ pixel frames). These tokens are decoded back to pixel space to obtain the coarse AR prediction $\hat{x}_{k+1},\hat{x}_{k+2},\hat{x}_{k+T}$. We employ a refinement module to further improve the prediction. It refines $\hat{x}_{j}$, the prediction of the AR model at time step $j$, to obtain a refined prediction $\hat{y}_{j}$. We incorporate the refinement module as we observe that the discretization process of the tokenizer introduces small rounding errors and leads to information loss. The refinement module, which has access to the historical context in pixel space, can mitigate such error by correcting the AR predictions based on the pixel values of the context. This process is visualized in Figure \ref{fig:explainer}.

\subsection{Universal Tokenizer}
We adopt the discrete tokenizer architecture from the Cosmos framework~\citep{agarwal2025cosmos}, which transforms input videos into sequences of compact discrete tokens while preserving spatiotemporal structure. The tokenizer follows an encoder-decoder design. The encoder processes an input sequence of video frames using temporally causal convolution and attention layers to produce a compressed latent representation. This representation is then quantized using Finite-Scalar Quantization (FSQ)~\citep{mentzerfinite}, resulting in discrete token sequences. The decoder reconstructs the original video frames from these tokens. A key design choice of the Cosmos tokenizer architecture is its temporal causality, where each stage observes only past and present frames, making it well-suited for downstream autoregressive modeling. In \name{}, we train a tokenizer with $8\times$ spatial and $4\times$ temporal compression. Given an input sequence of frames $x_{1:M}$, where $M=k+T$ and each frame has a size $H\times W$, the tokenizer then compresses them into $\hat{M}=\frac{k+T}{4}$ latent frames, each containing $L=\frac{HW}{8^2}$ discrete tokens. We concatenate the sequences to obtain a 1D sequence $z$.

To enable cross-task generalization, we train a single universal tokenizer across all available simulation datasets. This setting poses unique challenges due to the heterogeneity of the data: different datasets vary in channel dimensionality, spatial resolution, and physical semantics. We address this in two ways. First, we modify the first embedding layer of the encoder to accept a fixed union of all possible channels observed across datasets. When a data point lacks certain channels, we pad the missing entries with learnable 2D tensors specific to each channel type. This design allows the model to flexibly process any subset of channels within a unified architecture. Second, while the encoder is shared across datasets to enforce a shared embedding space, we train a separate decoder for each dataset to improve reconstruction quality and accommodate dataset-specific output distributions.

To ensure balanced representation across datasets during training, we replicate samples from datasets with fewer training examples so that each dataset contributes an equal number of sequences to the training process. We initialize the universal tokenizer from a pre-trained Cosmos checkpoint, which we found significantly accelerates convergence and improves reconstruction performance compared to training from scratch. This pre-trained initialization also facilitates better transfer to the physics domain by leveraging learned priors from natural video data.

\subsection{Autoregressive Generative Models}
After training the universal tokenizer, we train a large-scale autoregressive model to simulate physics in the discrete latent space. We adopt the autoregressive architecture introduced in the Cosmos framework~\citep{agarwal2025cosmos}, which is a decoder-only transformer trained using a next-token prediction objective. Given a sequence of discrete tokens produced by the tokenizer for the past $k$ input frames, the transformer predicts the tokens corresponding to the next $T$ frames autoregressively. The model minimizes the negative log-likelihood of the correct token at each position, conditioned on all previous tokens. Formally, the training objective is 
\begin{equation}
    \mathcal{L_{AR}}=-\sum _{i=1}^{\hat{M}}\sum _{j=1}^L \mathbb{E}_z\left[\log p(z_i^j|\{z_m^n|m<i \text{ or } m=i,n<j\})\right],
\end{equation}
where $L=\frac{HW}{8^2}$ is the length of each latent frame $z_i$, and $\hat{M}=\frac{k+T}{4}$ since each latent frame represents four pixel frames.

The autoregressive model incorporates 3D rotary position embeddings (RoPE) to capture relative spatiotemporal relationships across the token sequence. A key distinction from prior work is our support for variable spatial resolutions during training. Since simulation datasets differ in shape, we adjust the positional encodings dynamically: rather than resizing inputs or interpolating embeddings, we simply truncate the 3D RoPE frequencies along the height and width dimensions to match the size of the current input. This approach, implemented with minimal modification to the original RoPE module, allows seamless handling of mixed-resolution data without sacrificing performance. We found this simple strategy worked equally well as more advanced interpolation techniques~\citep{pengyarn,zhuolumina}.

We initialize the autoregressive model from the 4.5B parameter Cosmos checkpoint (\textsc{nvidia/Cosmos-1.0-Autoregressive-4B}), enabling it to inherit strong spatiotemporal priors learned from large-scale natural video datasets. Similar to tokenizer training, we oversample smaller datasets to match the size of the largest one.


\subsection{Refinement Module}
The refinement module is a convolutional neural network that aims to refine the output of AR models by removing the artifacts caused by the discretization process. We show one such example in Figure \ref{fig:explainer}. We observe that the output of the AR model (middle) $\hat{x}$ exhibits a pattern that is similar to quantization noise in the center, whereas the ground truth data (bottom) $x$ is noise-free. The refinement module is able to successfully remove such noise, as shown in its output (top) $\hat{y}$. This noise is introduced due to the inherent limitation of the discrete tokenization process, which was initially designed for natural videos. When generating natural videos such as characters or scenery, this noise is often negligible and does not affect the overall fidelity of generated videos. However, it can significantly hurt the performance in physical simulation tasks, where precision is required. 

We train our refinement module as a post-processing step for the AR model. After the AR model is trained, we autoregressively generate predictions for each sample in the training split to produce training data for the refinement module. The ground truth frames are used as the refinement target. Notably, we decode the outputs of the AR model before passing them into the refinement model, so the model learns to improve AR generations in pixel space. We adopt the same architecture as ConvNeXt-U-Net baseline of the Well benchmark for our refinement model and utilize MSE loss during the training process. The primary difference with the baseline is the learning objective, as the refinement model learns to refine the AR output instead of predicting a new frame itself. Just as our universal tokenizer employs different decoder layers for different datasets, we train separate refinement modules for each dataset. We provide more details in the appendix.
\begin{table}[b]
    \caption{\textbf{Next-frame prediction performance across 8 datasets on the Well benchmark}. We report VRMSE (lower is better) averaged across different fields for each dataset. }
    \label{tab:one_frame_prediction}
    \centering
    \small
    \setlength{\tabcolsep}{0.8em}
    \begin{tabular}{@{}l rrrr r@{}}
        \toprule
        \multirow{2}{*}{\texttt{Dataset}} 
          & \multicolumn{4}{c}{\textbf{Baseline}} 
          & \textbf{Ours} \\
        \cmidrule(l){2-5} \cmidrule(l){6-6}
          & FNO & TFNO & U-Net & C-U-Net 
          & \makecell{\name{}} \\
        \midrule
        \texttt{shear\_flow}                     
          & 1.189 & 1.472 & 3.447 & 0.8080 
          & \textbf{0.0700} \\
        \texttt{rayleigh\_benard}                
          & 0.8395 & 0.6566 & 1.4860 & 0.6699 
          & \textbf{0.1470} \\
        
        \texttt{acoustic\_scattering (maze)}     
          & 0.5062 & 0.5057 & 0.0351 & \textbf{0.0153}  
          & 0.0960 \\

        \texttt{active\_matter}                  
          & 0.3691 & 0.3598 & 0.2489 & 0.1034 
          & \textbf{0.0904} \\
        \texttt{turbulent\_radiative\_layer\_2D} 
          & 0.5001 & 0.5016 & 0.2418 & \textbf{0.1956} 
          & 0.2098 \\
        \texttt{viscoelastic\_instability}       
          & 0.7212 & 0.7102 & 0.4185 & 0.2499 
          & \textbf{0.2370} \\
        \texttt{gray\_scott\_reaction\_diffusion} 
          & 0.1365 & 0.3633 & 0.2252 & 0.1761 
          & \textbf{0.0210} \\
        \texttt{helmholtz\_staircase}            
          & \textbf{0.00046} & 0.00346 & 0.01931 & 0.02758
          & 0.0180 \\
        \midrule
        Average Rank ($\downarrow$) 
          & 3.62
          & 3.75
          & 3.62
          & 2.38
          & \textbf{1.62} \\
        \bottomrule
    \end{tabular}
\end{table}

\section{Experiments}
We train and evaluate \name{} across eight simulation tasks from the Well benchmark~\citep{ohana2024well}, as shown in Tables~\ref{tab:one_frame_prediction} and~\ref{tab:rollout_table}. Following the benchmark protocol, we report the Variance-Weighted Root Mean Squared Error (VRMSE), averaged over all physical channels for each dataset. For datasets such as \texttt{helmholtz\_staircase} and \texttt{acoustic\_scattering (maze)}, we exclude channels that remain constant across time steps from the evaluation. We compare \name{} against four baselines provided by the Well benchmark: Fourier Neural Operator (FNO), Tucker-Factorized FNO (TFNO), U-Net, and U-Net with ConvNeXt blocks (C-U-Net), considering both next-frame and long-horizon rollout settings. In addition, we conduct extensive ablation studies to assess the impact of various architectural and training design choices in \name{}.

\subsection{Next-frame Prediction}
In the next-frame prediction benchmark, \name{} outperforms the baselines on $5$ out of $8$ datasets, demonstrating strong generalization across diverse physical systems. In addition, \name{} achieves the best average rank across the 8 tasks, with a score of $1.62$ compared to $2.38$ for the best-performing baseline. Importantly, \name{} achieves this performance using a single model checkpoint shared across all tasks, whereas the baseline results are obtained from separate models trained specifically for each dataset. This highlights the ability of \name{} to act as a general-purpose simulator. The performance gain is especially significant on the \texttt{shear\_flow} and \texttt{rayleigh\_benard} datasets, where \name{} reduces the VRMSE by $91\%$ and $78\%$ respectively relative to the best baseline.

\begin{table}[tbp]
  \caption{\textbf{Long-horizon prediction performance across 8 datasets on the Well benchmark.} We report VRMSE (lower is better) averaged across different fields for each dataset. We report averaged results over different ranges of lead time: 2-8, 9-26 and 27-56 frames.}
  \label{tab:rollout_table}
  \centering
  \small
  \begin{tabular}{@{}l cc cc cc@{}}
    \toprule
    \multirow{2}{*}{\texttt{Dataset}}
      & \multicolumn{2}{c}{$\Delta t$ =2:8}
      & \multicolumn{2}{c}{$\Delta t$ =9:26}
      & \multicolumn{2}{c}{$\Delta t$ =27:56} \\
    \cmidrule(lr){2-3} \cmidrule(lr){4-5} \cmidrule(lr){6-7}
      & Baseline & \name{}
      & Baseline & \name{}
      & Baseline & \name{} \\
    \midrule
    \texttt{shear\_flow}
      & 2.330    & \textbf{0.077}
      & $>$10    & \textbf{0.153}
      & $>$10    & \textbf{0.236} \\

    \texttt{rayleigh\_benard}
      & $>$10    & \textbf{1.067}
      & $>$10    & \textbf{0.741}
      & $>$10    & \textbf{0.847} \\

    \texttt{acoustic\_scattering (maze)}
      & 0.560    & \textbf{0.158}
      & \textbf{0.920}    & 1.246
      & \textbf{1.341}    & 2.189 \\

    \texttt{active\_matter}
      & 2.110    & \textbf{0.415}
      & 2.710    & \textbf{0.974}
      & 1.635    & \textbf{1.320} \\

    \texttt{turbulent\_radiative\_layer\_2D}
      & 0.660    & \textbf{0.363}
      & 1.040    & \textbf{0.693}
      & 1.331    & \textbf{0.953} \\

    \texttt{gray\_scott\_reaction\_diffusion}
      & 0.290    & \textbf{0.037}
      & 7.620    & \textbf{1.984}
      & 12.714   & \textbf{12.643} \\

    \texttt{viscoelastic\_instability}
      & 0.520    & \textbf{0.387}
      & —        & —
      & —        & — \\

    \texttt{helmholtz\_staircase}
      & \textbf{0.002} & 0.022
      & \textbf{0.003} & 0.071
      & —        & — \\
    \bottomrule
  \end{tabular}
\end{table}
\begin{figure}
    \centering
    \includegraphics[width=1\linewidth]{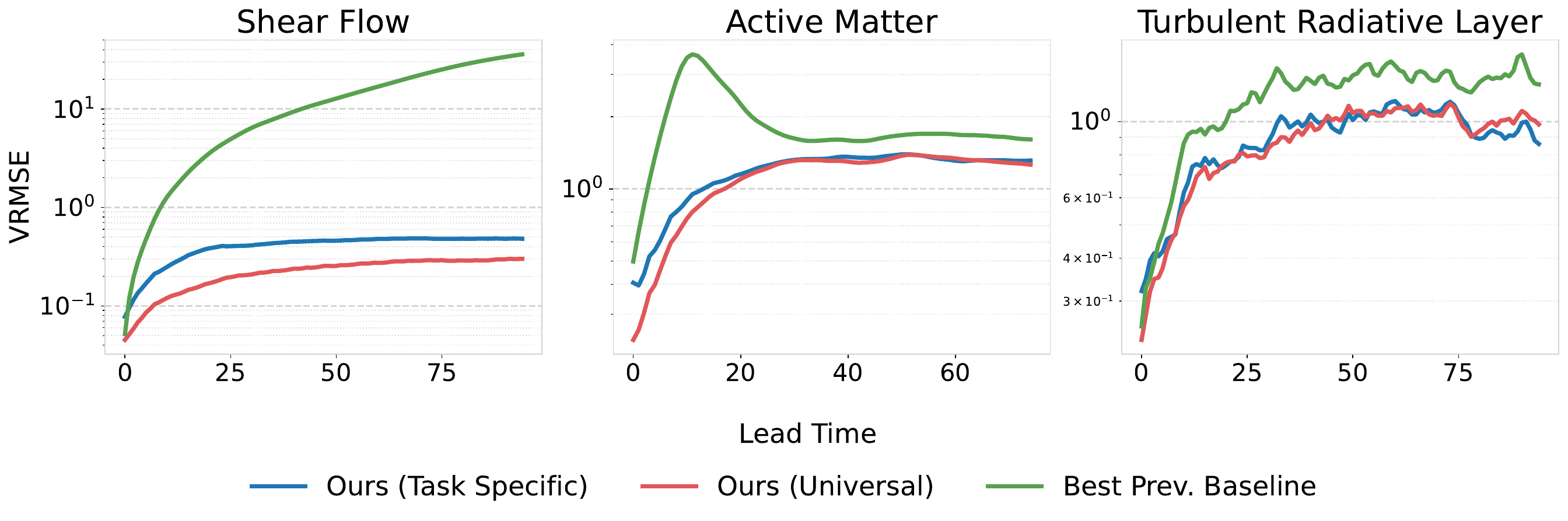}
    \caption{\textbf{Long-horizon prediction performance.} We visualize VRMSE (lower is better) across different lead time on \texttt{shear\_flow},\texttt{active\_matter}, and \texttt{turbulent\_radiative\_layer} datasets.  }
    \label{fig:rollout}
\end{figure}

\subsection{Long-horizon Prediction}
While \name{} already performs competitively in next-frame prediction, its true strength lies in long-horizon simulation. 
As shown in Table~\ref{tab:rollout_table}, \name{} achieves state-of-the-art performance on $18/21$ evaluation points across different forecasting windows. The improvements are not only consistent but also significant in various tasks. For example, on \texttt{shear\_flow}, \name{} reduces VRMSE by over $97\%$ at the $6$:$12$ horizon compared to the best-performing baseline (from $2.33$ to $0.077$). On \texttt{rayleigh\_benard}, \name{} achieves more than $90\%$ lower error across all rollout windows. Similar results are observed in \texttt{active\_matter}, where \name{} consistently achieves better performance at every forecast horizon, underscoring its robustness and adaptability across domains.

Figure~\ref{fig:rollout} further illustrates the long-term behavior of \name{} compared to the best baseline model in each dataset. In the early stages of the rollout, both models exhibit similar performance. However, as the lead time increases, the performance of the baseline models degrades rapidly due to compounding prediction errors. In contrast, \name{} maintains low VRMSE across time steps, demonstrating much greater stability. 
This stability stems from the autoregressive nature of \name{}, which allows the model to learn from full sequences of simulations rather than focusing solely on short-term prediction. This enables it to maintain stability and accuracy over extended rollouts, making it particularly well-suited for challenging multi-step prediction tasks.

\subsection{Ablation Studies}

To study the effectiveness of our design, we conducted a series of thorough ablation studies. In the main paper, we explored the performance of universal (multi-task) models versus single-task models, and the effectiveness of the refinement module. We provide additional ablation studies, such as training the model from scratch versus initializing the model with weights pre-trained on natural videos in the appendix. 

\begin{table}[t]
  \centering
  \small
    \caption{\textbf{Comparison of multi- and single-task models.} We report next-frame and long-horizon prediction results on the Well benchmark for the multi-task and single-task models.}
  \label{tab:ablation-specialize}
  \begin{tabular}{@{}l cc cc cc cc@{}}
    \toprule
    \multirow{2}{*}{\texttt{Dataset}}
      &  \multicolumn{2}{c}{$\Delta t$ =1}
      &  \multicolumn{2}{c}{$\Delta t$ =2:8}
      &  \multicolumn{2}{c}{$\Delta t$ =9:26}
      &  \multicolumn{2}{c}{$\Delta t$ =27:56} \\
    \cmidrule(lr){2-3} \cmidrule(lr){4-5} \cmidrule(lr){6-7} \cmidrule(lr){8-9}
      & Spec. & Univ.
      & Spec. & Univ.
      & Spec. & Univ.
      & Spec. & Univ. \\
    \midrule
    \texttt{shear\_flow}
      & \textbf{0.0689}  & 0.070
      & 0.236           & \textbf{0.118}
      & 0.378           & \textbf{0.281}
      & 0.452           & \textbf{0.397} \\

    \texttt{rayleigh\_benard}
      & \textbf{0.137}   & 0.147
      & \textbf{0.436} & 1.090
      & \textbf{0.522} & 0.704
      & 0.724           & \textbf{0.646} \\

    \texttt{turbulent\_radiative\_layer}
      & 0.359            & \textbf{0.343}
      & 0.565           & \textbf{0.357}
      & 0.792           & \textbf{0.710}
      & 1.014           & \textbf{0.998} \\

    \texttt{active\_matter}
      & 0.150            & \textbf{0.090}
      & 0.844           & \textbf{0.477}
      & \textbf{1.177} & 1.396
      & \textbf{1.352} & 1.381 \\

    \texttt{gray\_scott\_reaction}
      & 0.0418           & \textbf{0.0210}
      & 1.487           & \textbf{0.0375}
      & 15.965          & \textbf{0.390}
      & 62.484          & \textbf{0.895} \\

    \texttt{viscoelastic\_instability}
      & 0.251            & \textbf{0.237}
      & 0.764           & \textbf{0.406}
      & —               & —
      & —               & — \\
    \bottomrule
  \end{tabular}
\end{table}
\begin{figure}[t]
    \centering
    \includegraphics[width=1\linewidth]{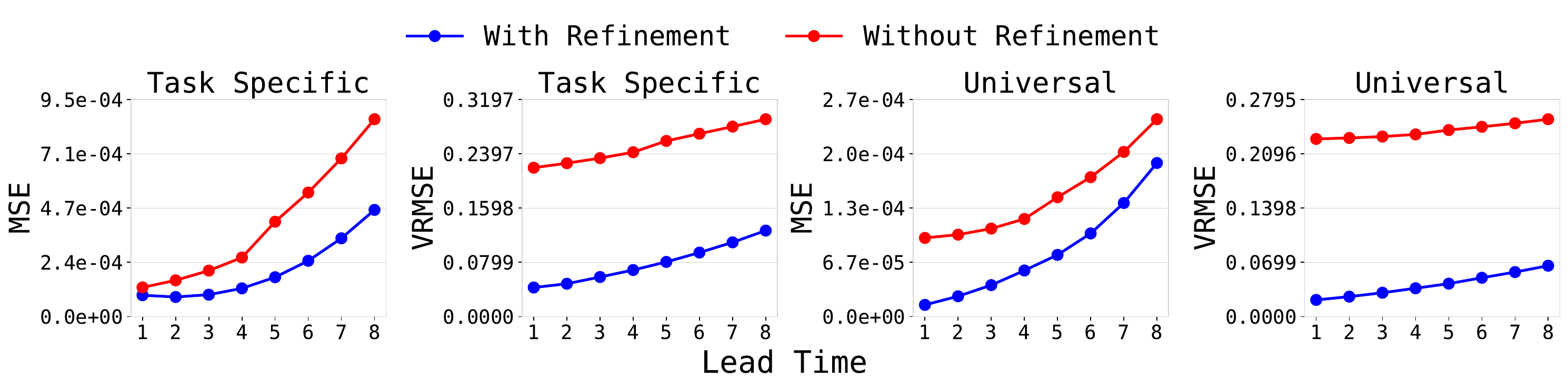}
    \caption{\textbf{Effect of refinement module.} We apply refinement module to both the multi-task and single-task AR model and study its effect on predication errors. We report VRMSE and MSE (lower is better) over prediction windows ranging from 1 frame to 8 frames on the \texttt{gray\_scott\_reaction\_diffusion} dataset. }
    \label{fig:refiner-gs}
\end{figure}

\textbf{General Model vs Task Specific Models }
We compare the performance of our multi-task model and single-task models on both one-frame prediction and long-horizon prediction tasks. For the task-specific model, we followed the same setup as the universal model, including the model size, model architecture, and training hyperparameters. The only difference is the training data.  We report VRMSE across 8 datasets and different lead times in Table \ref{tab:ablation-specialize}. Experiment results show that the universal model outperforms task-specific models, achieving lower VRMSE on the majority of datasets across different lead times. Our results show that joint multi-task training improves the performance of individual tasks, as the model may learn some common patterns and mechanisms across different physical processes.

\textbf{Effectiveness of Refinement Module }
We compare \name{}~with and without the refinement module. We show such differences for both the multi-task AR model and the single-task AR model at different prediction windows in Figure \ref{fig:refiner-gs}. The refinement model reduces MSE and VRMSE metrics for both models on all prediction windows of the \texttt{gray\_scott\_reaction\_diffusion} dataset, highlighting the effectiveness of the proposed refinement process. Most notably, with the help of refinement model, the 8-frame prediction error (0.07) of our multi-task model, measured by VRMSE, is lower than the 1-frame prediction error of the best performing baseline on the Well benchmark (0.14).

\subsection{Adaptation to Unseen Simulations} \label{sec:adaptation}
We evaluate the adaptability of \name{} on two unseen simulations: \texttt{euler\_multi\_quadrants (periodic b.c.)} and \texttt{acoustic\_scattering (discontinuous)}. These tasks involve novel input channels and physical dynamics not encountered during training. To handle this distribution shift, we fully finetune the tokenizer for each task. We consider two variants of the autoregressive model: \name{}$_f$, which finetunes the pretrained model, and \name{}$_s$, which trains from scratch using the Cosmos checkpoint as initialization. Further finetuning details are provided in Appendix~\ref{sec:experimental-settings}.

Table~\ref{tab:finetuning_results} shows that \name{}$_f$ achieves the best performance on nearly all tasks and prediction horizons, only losing to C-U-Net on one-step prediction for one task, and the performance gap widens significantly as the horizon increases. Notably, \name{}$_f$ consistently outperforms \name{}$_s$ across all settings, highlighting its ability to effectively transfer knowledge to previously unseen simulations.

\begin{table}[t]
  \centering
  \small
    \caption{\textbf{Performance on two simulation tasks unseen during training.} We compare both the finetuning version (\name{}$_f$) and the scratch version (\name{}$_s$) with the baselines.}
  \label{tab:finetuning_results}
  \resizebox{1.0\textwidth}{!}{
  \begin{tabular}{@{}l cc cc cc cc@{}}
    \toprule
    \multirow{2}{*}{\texttt{Models}}
      &  \multicolumn{4}{c}{\texttt{euler\_multi\_quadrants (periodic b.c.)}}
      &  \multicolumn{4}{c}{\texttt{acoustic\_scattering (discontinuous)}} \\
    \cmidrule(lr){2-5} \cmidrule(lr){6-9}
      & $\Delta t = 1$ & $\Delta t = $2:8 & $\Delta t = $9:26 & $\Delta t = $27:56
      & $\Delta t = 1$ & $\Delta t = $2:8 & $\Delta t = $9:26 & $\Delta t = $27:56 \\
    \midrule
    \name{}$_f$
      & \textbf{0.105} & \textbf{0.188} & \textbf{0.358} & \textbf{0.642} & 0.038 & \textbf{0.057} & \textbf{0.443} & \textbf{1.168}  \\

    \name{}$_s$
      & \textbf{0.105} & \textbf{0.188} & 0.366 & 0.658 & 0.039 & 0.062 & 0.455 & 1.192  \\

    FNO
      & 0.408 & 1.130 & 1.370 & -- & 0.127 & 2.146 & 2.752 & 3.135  \\

    TFNO
      & 0.416 & 1.230 & 1.520 & -- & 0.130 & 2.963 & 3.713 & 4.081  \\

    U-Net
      & 0.183 & 1.020 & 1.630 & -- & 0.045 & 2.855 & 6.259 & 8.074  \\

    C-U-Net
      & 0.153 & 4.980 & $>$10 & -- & \textbf{0.006} & 5.160 & $>$10 & $>$10  \\

    \bottomrule
  \end{tabular}
  }
\end{table}

\begin{figure}[h!]
    \centering
    \includegraphics[width=0.8\linewidth]{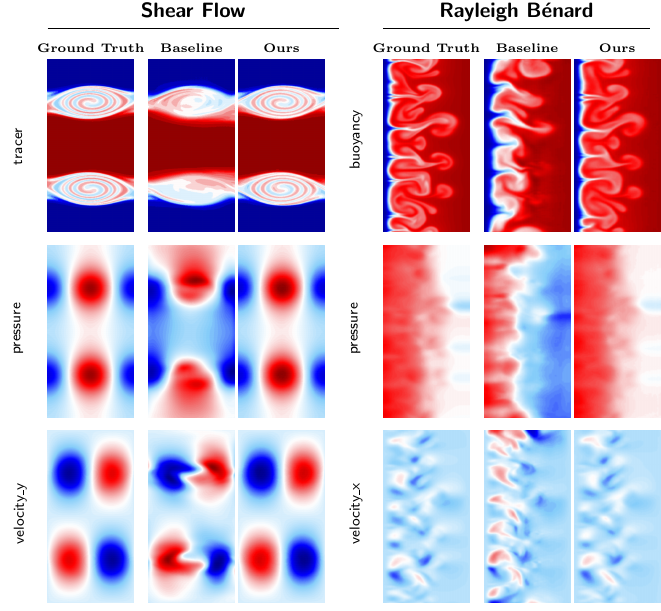}
    \caption{\textbf{Side-by-side qualitative comparison of \name{} and baseline models.} \name{} demonstrates superior performance in long horizon rollouts than the leading baseline model. At lead times of 24 and 15 steps for shear flow and Rayleigh–Bénard convection respectively, \name{} maintains high-fidelity predictions across all physical fields, while baseline models ConvNeXt-UNet and TFNO exhibit visible distortions and loss of detail.}
    \label{fig:qualitative_samples}
\end{figure}

\subsection{Qualitative Comparison}
Figure~\ref{fig:qualitative_samples} presents a qualitative comparison between \name{} and the best-performing baseline models on two representative simulation tasks: \texttt{shear\_flow} and \texttt{rayleigh\_benard}. At rollout horizons of $24$ and $15$ steps respectively, \name{} produces predictions that remain visually consistent with the ground truth across all physical fields, including tracer, pressure, buoyancy, and velocity components. In contrast, baseline models exhibit noticeable distortions, blurring, and loss of fine-grained structures, particularly evident in the vortex structures of \texttt{shear\_flow} and the convective plumes of \texttt{rayleigh\_benard}. These qualitative results highlight superior fidelity and stability of \name{} over extended prediction windows.

\section{Conclusion}
\name{} introduces a unified foundational model designed for general-purpose physical simulation across a diverse range of systems. \name{} uses a universal tokenizer for shared discrete representations, a large-scale autoregressive transformer for modeling temporal relationships, and a refinement network to improve output fidelity. Our joint training approach enabled \name{} to capture shared spatiotemporal patterns and adapt to varying resolutions, channel configurations, and physical semantics. We show a single universally trained model significantly outperforms task-specific baselines on a wide variety of physical domains. \name{} demonstrates superior performance on single timestep prediction and significantly outperforms baseline methods on long-horizon rollouts, while maintaining stability and accuracy. The success of \name{} highlights the potential of foundation models in accelerating scientific discovery by providing generalizable tools to model complex physical phenomena. 

\section{Acknowledgement}
AG would like to acknowledge support from NSF CAREER Grant \#2341040, Schmidt Sciences Early Career Fellowship, and Amazon Research Award.

\clearpage
\bibliographystyle{plain}
\bibliography{ref}

\begin{thebibliography}{10}

\bibitem{abu2016youtube}
Sami Abu-El-Haija, Nisarg Kothari, Joonseok Lee, Paul Natsev, George Toderici, Balakrishnan Varadarajan, and Sudheendra Vijayanarasimhan.
\newblock Youtube-8m: A large-scale video classification benchmark.
\newblock {\em arXiv preprint arXiv:1609.08675}, 2016.

\bibitem{agarwal2025cosmos}
Niket Agarwal, Arslan Ali, Maciej Bala, Yogesh Balaji, Erik Barker, Tiffany Cai, Prithvijit Chattopadhyay, Yongxin Chen, Yin Cui, Yifan Ding, et~al.
\newblock Cosmos world foundation model platform for physical ai.
\newblock {\em arXiv preprint arXiv:2501.03575}, 2025.

\bibitem{bain2021frozen}
Max Bain, Arsha Nagrani, G{\"u}l Varol, and Andrew Zisserman.
\newblock Frozen in time: A joint video and image encoder for end-to-end retrieval.
\newblock In {\em Proceedings of the IEEE/CVF international conference on computer vision}, pages 1728--1738, 2021.

\bibitem{bar2024lumiere}
Omer Bar-Tal, Hila Chefer, Omer Tov, Charles Herrmann, Roni Paiss, Shiran Zada, Ariel Ephrat, Junhwa Hur, Guanghui Liu, Amit Raj, et~al.
\newblock Lumiere: A space-time diffusion model for video generation.
\newblock In {\em SIGGRAPH Asia 2024 Conference Papers}, pages 1--11, 2024.

\bibitem{berger2024implicit}
Marsha~J Berger and Randall~J LeVeque.
\newblock Implicit adaptive mesh refinement for dispersive tsunami propagation.
\newblock {\em SIAM Journal on Scientific Computing}, 46(4):B554--B578, 2024.

\bibitem{biegler2003large}
Lorenz~T Biegler, Omar Ghattas, Matthias Heinkenschloss, and Bart van Bloemen~Waanders.
\newblock Large-scale pde-constrained optimization: an introduction.
\newblock In {\em Large-scale PDE-constrained optimization}, pages 3--13. Springer, 2003.

\bibitem{bommasani2021opportunities}
Rishi Bommasani, Drew~A Hudson, Ehsan Adeli, Russ Altman, Simran Arora, Sydney von Arx, Michael~S Bernstein, Jeannette Bohg, Antoine Bosselut, Emma Brunskill, et~al.
\newblock On the opportunities and risks of foundation models.
\newblock {\em arXiv preprint arXiv:2108.07258}, 2021.

\bibitem{brandstetter2022message}
Johannes Brandstetter, Daniel Worrall, and Max Welling.
\newblock Message passing neural pde solvers.
\newblock {\em arXiv preprint arXiv:2202.03376}, 2022.

\bibitem{brown2020language}
Tom Brown, Benjamin Mann, Nick Ryder, Melanie Subbiah, Jared~D Kaplan, Prafulla Dhariwal, Arvind Neelakantan, Pranav Shyam, Girish Sastry, Amanda Askell, et~al.
\newblock Language models are few-shot learners.
\newblock {\em Advances in neural information processing systems}, 33:1877--1901, 2020.

\bibitem{chen2020simple}
Ting Chen, Simon Kornblith, Mohammad Norouzi, and Geoffrey Hinton.
\newblock A simple framework for contrastive learning of visual representations.
\newblock In {\em International conference on machine learning}, pages 1597--1607. PmLR, 2020.

\bibitem{choudhary2022recent}
Kamal Choudhary, Brian DeCost, Chi Chen, Anubhav Jain, Francesca Tavazza, Ryan Cohn, Cheol~Woo Park, Alok Choudhary, Ankit Agrawal, Simon~JL Billinge, et~al.
\newblock Recent advances and applications of deep learning methods in materials science.
\newblock {\em npj Computational Materials}, 8(1):59, 2022.

\bibitem{cranmer2020frontier}
Kyle Cranmer, Johann Brehmer, and Gilles Louppe.
\newblock The frontier of simulation-based inference.
\newblock {\em Proceedings of the National Academy of Sciences}, 117(48):30055--30062, 2020.

\bibitem{eyring2016overview}
Veronika Eyring, Sandrine Bony, Gerald~A Meehl, Catherine~A Senior, Bjorn Stevens, Ronald~J Stouffer, and Karl~E Taylor.
\newblock Overview of the coupled model intercomparison project phase 6 (cmip6) experimental design and organization.
\newblock {\em Geoscientific Model Development}, 9(5):1937--1958, 2016.

\bibitem{franco2023deep}
Nicola~Rares Franco, Stefania Fresca, Filippo Tombari, and Andrea Manzoni.
\newblock Deep learning-based surrogate models for parametrized pdes: Handling geometric variability through graph neural networks.
\newblock {\em Chaos: An Interdisciplinary Journal of Nonlinear Science}, 33(12), 2023.

\bibitem{gafni2022make}
Oran Gafni, Adam Polyak, Oron Ashual, Shelly Sheynin, Devi Parikh, and Yaniv Taigman.
\newblock Make-a-scene: Scene-based text-to-image generation with human priors.
\newblock In {\em European Conference on Computer Vision}, pages 89--106. Springer, 2022.

\bibitem{goldberg2022numerical}
Jared~A Goldberg, Yan-Fei Jiang, and Lars Bildsten.
\newblock Numerical simulations of convective three-dimensional red supergiant envelopes.
\newblock {\em The Astrophysical Journal}, 929(2):156, 2022.

\bibitem{gopakumar2024plasma}
Vignesh Gopakumar, Stanislas Pamela, Lorenzo Zanisi, Zongyi Li, Ander Gray, Daniel Brennand, Nitesh Bhatia, Gregory Stathopoulos, Matt Kusner, Marc~Peter Deisenroth, et~al.
\newblock Plasma surrogate modelling using fourier neural operators.
\newblock {\em Nuclear Fusion}, 64(5):056025, 2024.

\bibitem{gupta2022towards}
Jayesh~K Gupta and Johannes Brandstetter.
\newblock Towards multi-spatiotemporal-scale generalized pde modeling.
\newblock {\em arXiv preprint arXiv:2209.15616}, 2022.

\bibitem{haghighat2021physics}
Ehsan Haghighat, Maziar Raissi, Adrian Moure, Hector Gomez, and Ruben Juanes.
\newblock A physics-informed deep learning framework for inversion and surrogate modeling in solid mechanics.
\newblock {\em Computer Methods in Applied Mechanics and Engineering}, 379:113741, 2021.

\bibitem{he2020momentum}
Kaiming He, Haoqi Fan, Yuxin Wu, Saining Xie, and Ross Girshick.
\newblock Momentum contrast for unsupervised visual representation learning.
\newblock In {\em Proceedings of the IEEE/CVF conference on computer vision and pattern recognition}, pages 9729--9738, 2020.

\bibitem{ho2022video}
Jonathan Ho, Tim Salimans, Alexey Gritsenko, William Chan, Mohammad Norouzi, and David~J Fleet.
\newblock Video diffusion models.
\newblock {\em Advances in Neural Information Processing Systems}, 35:8633--8646, 2022.

\bibitem{karniadakis2021physics_informed_review}
George~Em Karniadakis, Ioannis~G Kevrekidis, Lu~Lu, Paris Perdikaris, Sifan Wang, and Liu Yang.
\newblock Physics-informed machine learning.
\newblock {\em Nature Reviews Physics}, 3(6):422--440, 2021.

\bibitem{kim2022transfer}
Hee~E Kim, Alejandro Cosa-Linan, Nandhini Santhanam, Mahboubeh Jannesari, Mate~E Maros, and Thomas Ganslandt.
\newblock Transfer learning for medical image classification: a literature review.
\newblock {\em BMC medical imaging}, 22(1):69, 2022.

\bibitem{kingma2014adam}
Diederik~P Kingma and Jimmy Ba.
\newblock Adam: A method for stochastic optimization.
\newblock {\em arXiv preprint arXiv:1412.6980}, 2014.

\bibitem{kissas2022learning}
Georgios Kissas, Jacob~H Seidman, Leonardo~Ferreira Guilhoto, Victor~M Preciado, George~J Pappas, and Paris Perdikaris.
\newblock Learning operators with coupled attention.
\newblock {\em Journal of Machine Learning Research}, 23(215):1--63, 2022.

\bibitem{kondratyuk2023videopoet}
Dan Kondratyuk, Lijun Yu, Xiuye Gu, Jos{\'e} Lezama, Jonathan Huang, Grant Schindler, Rachel Hornung, Vighnesh Birodkar, Jimmy Yan, Ming-Chang Chiu, et~al.
\newblock Videopoet: A large language model for zero-shot video generation.
\newblock {\em arXiv preprint arXiv:2312.14125}, 2023.

\bibitem{kong2024hunyuanvideo}
Weijie Kong, Qi~Tian, Zijian Zhang, Rox Min, Zuozhuo Dai, Jin Zhou, Jiangfeng Xiong, Xin Li, Bo~Wu, Jianwei Zhang, et~al.
\newblock Hunyuanvideo: A systematic framework for large video generative models.
\newblock {\em arXiv preprint arXiv:2412.03603}, 2024.

\bibitem{kovachki2023neural}
Nikola~B. Kovachki, Zongyi Li, Burigede Liu, Kamyar Azizzadenesheli, Kaushik Bhattacharya, Andrew~M. Stuart, and Anima Anandkumar.
\newblock Neural operator: Learning maps between function spaces.
\newblock {\em Journal of Machine Learning Research}, 24:1--97, 2023.
\newblock Article 89.

\bibitem{lemos2023simbig}
Pablo Lemos, Liam Parker, ChangHoon Hahn, Shirley Ho, Michael Eickenberg, Jiamin Hou, Elena Massara, Chirag Modi, Azadeh~Moradinezhad Dizgah, Bruno Regaldo-Saint Blancard, et~al.
\newblock Simbig: Field-level simulation-based inference of galaxy clustering.
\newblock {\em arXiv preprint arXiv:2310.15256}, 2023.

\bibitem{li2022transformer}
Zijie Li, Kazem Meidani, and Amir~Barati Farimani.
\newblock Transformer for partial differential equations' operator learning.
\newblock {\em arXiv preprint arXiv:2205.13671}, 2022.

\bibitem{li2020neural}
Zongyi Li, Nikola Kovachki, Kamyar Azizzadenesheli, Burigede Liu, Kaushik Bhattacharya, Andrew Stuart, and Anima Anandkumar.
\newblock Neural operator: Graph kernel network for partial differential equations.
\newblock {\em arXiv preprint arXiv:2003.03485}, 2020.

\bibitem{li2021fourier}
Zongyi Li, Nikola Kovachki, Kamyar Azizzadenesheli, Burigede Liu, Kaushik Bhattacharya, Andrew Stuart, and Anima Anandkumar.
\newblock Fourier neural operator for parametric partial differential equations.
\newblock In {\em International Conference on Learning Representations}. ICLR, 2021.

\bibitem{lippe2023pde}
Phillip Lippe, Bas Veeling, Paris Perdikaris, Richard Turner, and Johannes Brandstetter.
\newblock Pde-refiner: Achieving accurate long rollouts with neural pde solvers.
\newblock {\em Advances in Neural Information Processing Systems}, 36:67398--67433, 2023.

\bibitem{lu2019deeponet}
Lu~Lu, Pengzhan Jin, and George~Em Karniadakis.
\newblock Deeponet: Learning nonlinear operators for identifying differential equations based on the universal approximation theorem of operators.
\newblock {\em arXiv preprint arXiv:1910.03193}, 2019.

\bibitem{lu2021learning}
Lu~Lu, Pengzhan Jin, Guofei Pang, Zhongqiang Zhang, and George~Em Karniadakis.
\newblock Learning nonlinear operators via {DeepONet} based on the universal approximation theorem of operators.
\newblock {\em Nature Machine Intelligence}, 3(3):218--229, 2021.

\bibitem{mentzerfinite}
Fabian Mentzer, David Minnen, Eirikur Agustsson, and Michael Tschannen.
\newblock Finite scalar quantization: Vq-vae made simple.
\newblock In {\em The Twelfth International Conference on Learning Representations}.

\bibitem{mohammadi2004shape}
Bijan Mohammadi and Olivier Pironneau.
\newblock Shape optimization in fluid mechanics.
\newblock {\em Annu. Rev. Fluid Mech.}, 36(1):255--279, 2004.

\bibitem{nguyen2024efficient}
Binh~Duong Nguyen, Pavlo Potapenko, Aytekin Demirci, Kishan Govind, S{\'e}bastien Bompas, and Stefan Sandfeld.
\newblock Efficient surrogate models for materials science simulations: Machine learning-based prediction of microstructure properties.
\newblock {\em Machine learning with applications}, 16:100544, 2024.

\bibitem{nguyen2023climax}
Tung Nguyen, Johannes Brandstetter, Ashish Kapoor, Jayesh~K Gupta, and Aditya Grover.
\newblock Climax: A foundation model for weather and climate.
\newblock {\em arXiv preprint arXiv:2301.10343}, 2023.

\bibitem{ohana2024well}
Ruben Ohana, Michael McCabe, Lucas Meyer, Rudy Morel, Fruzsina Agocs, Miguel Beneitez, Marsha Berger, Blakesly Burkhart, Stuart Dalziel, Drummond Fielding, et~al.
\newblock The well: a large-scale collection of diverse physics simulations for machine learning.
\newblock {\em Advances in Neural Information Processing Systems}, 37:44989--45037, 2024.

\bibitem{openai2024sora}
OpenAI.
\newblock Sora: A video generation model.
\newblock \url{https://openai.com/sora}, 2024.
\newblock Accessed: 2025-05-13.

\bibitem{oquab2023dinov2}
Maxime Oquab, Timoth{\'e}e Darcet, Th{\'e}o Moutakanni, Huy Vo, Marc Szafraniec, Vasil Khalidov, Pierre Fernandez, Daniel Haziza, Francisco Massa, Alaaeldin El-Nouby, et~al.
\newblock Dinov2: Learning robust visual features without supervision.
\newblock {\em arXiv preprint arXiv:2304.07193}, 2023.

\bibitem{pengyarn}
Bowen Peng, Jeffrey Quesnelle, Honglu Fan, and Enrico Shippole.
\newblock Yarn: Efficient context window extension of large language models.
\newblock In {\em The Twelfth International Conference on Learning Representations}.

\bibitem{qing2024hierarchical}
Zhiwu Qing, Shiwei Zhang, Jiayu Wang, Xiang Wang, Yujie Wei, Yingya Zhang, Changxin Gao, and Nong Sang.
\newblock Hierarchical spatio-temporal decoupling for text-to-video generation.
\newblock In {\em Proceedings of the IEEE/CVF Conference on Computer Vision and Pattern Recognition}, pages 6635--6645, 2024.

\bibitem{radford2021learning}
Alec Radford, Jong~Wook Kim, Chris Hallacy, Aditya Ramesh, Gabriel Goh, Sandhini Agarwal, Girish Sastry, Amanda Askell, Pamela Mishkin, Jack Clark, et~al.
\newblock Learning transferable visual models from natural language supervision.
\newblock In {\em International conference on machine learning}, pages 8748--8763. PmLR, 2021.

\bibitem{raissi2019physics}
Maziar Raissi, Paris Perdikaris, and George~Em Karniadakis.
\newblock Physics-informed neural networks: A deep learning framework for solving forward and inverse problems involving nonlinear partial differential equations.
\newblock {\em Journal of Computational Physics}, 378:686--707, 2019.

\bibitem{reed2023scale}
Colorado~J Reed, Ritwik Gupta, Shufan Li, Sarah Brockman, Christopher Funk, Brian Clipp, Kurt Keutzer, Salvatore Candido, Matt Uyttendaele, and Trevor Darrell.
\newblock Scale-mae: A scale-aware masked autoencoder for multiscale geospatial representation learning.
\newblock In {\em Proceedings of the IEEE/CVF International Conference on Computer Vision}, pages 4088--4099, 2023.

\bibitem{ronneberger2015u}
Olaf Ronneberger, Philipp Fischer, and Thomas Brox.
\newblock U-net: Convolutional networks for biomedical image segmentation.
\newblock In {\em International Conference on Medical image computing and computer-assisted intervention}, pages 234--241. Springer, 2015.

\bibitem{ryczko2019deep}
Kevin Ryczko, David~A Strubbe, and Isaac Tamblyn.
\newblock Deep learning and density-functional theory.
\newblock {\em Physical Review A}, 100(2):022512, 2019.

\bibitem{siahkoohi2023martian}
Ali Siahkoohi, Rudy Morel, Randall Balestriero, Erwan Allys, Gr{\'e}gory Sainton, Taichi Kawamura, and Maarten~V de~Hoop.
\newblock Martian time-series unraveled: A multi-scale nested approach with factorial variational autoencoders.
\newblock {\em arXiv preprint arXiv:2305.16189}, 2023.

\bibitem{subramanian2023towards}
Shashank Subramanian, Peter Harrington, Kurt Keutzer, Wahid Bhimji, Dmitriy Morozov, Michael~W. Mahoney, and Amir Gholami.
\newblock Towards foundation models for scientific machine learning: Characterizing scaling and transfer behavior.
\newblock In {\em Thirty-seventh Conference on Neural Information Processing Systems}, 2023.

\bibitem{sun2020surrogate}
Luning Sun, Han Gao, Shaowu Pan, and Jian-Xun Wang.
\newblock Surrogate modeling for fluid flows based on physics-constrained deep learning without simulation data.
\newblock {\em Computer Methods in Applied Mechanics and Engineering}, 361:112732, 2020.

\bibitem{sun2025ar}
Mingzhen Sun, Weining Wang, Gen Li, Jiawei Liu, Jiahui Sun, Wanquan Feng, Shanshan Lao, SiYu Zhou, Qian He, and Jing Liu.
\newblock Ar-diffusion: Asynchronous video generation with auto-regressive diffusion.
\newblock {\em arXiv preprint arXiv:2503.07418}, 2025.

\bibitem{takeda2023multi}
Seiji Takeda, Indra Priyadarsini, Akihiro Kishimoto, Hajime Shinohara, Lisa Hamada, Hirose Masataka, Junta Fuchiwaki, and Daiju Nakano.
\newblock Multi-modal foundation model for material design.
\newblock In {\em AI for Accelerated Materials Design-NeurIPS 2023 Workshop}, 2023.

\bibitem{tao2019application}
Jun Tao and Gang Sun.
\newblock Application of deep learning based multi-fidelity surrogate model to robust aerodynamic design optimization.
\newblock {\em Aerospace Science and Technology}, 92:722--737, 2019.

\bibitem{torlai2018neural}
Giacomo Torlai, Guglielmo Mazzola, Juan Carrasquilla, Matthias Troyer, Roger Melko, and Giuseppe Carleo.
\newblock Neural-network quantum state tomography.
\newblock {\em Nature physics}, 14(5):447--450, 2018.

\bibitem{wang2025wan}
Ang Wang, Baole Ai, Bin Wen, Chaojie Mao, Chen-Wei Xie, Di~Chen, Feiwu Yu, Haiming Zhao, Jianxiao Yang, Jianyuan Zeng, et~al.
\newblock Wan: Open and advanced large-scale video generative models.
\newblock {\em arXiv preprint arXiv:2503.20314}, 2025.

\bibitem{wang2024omnitokenizer}
Junke Wang, Yi~Jiang, Zehuan Yuan, Bingyue Peng, Zuxuan Wu, and Yu-Gang Jiang.
\newblock Omnitokenizer: A joint image-video tokenizer for visual generation.
\newblock {\em Advances in Neural Information Processing Systems}, 37:28281--28295, 2024.

\bibitem{wang2024emu3}
Xinlong Wang, Xiaosong Zhang, Zhengxiong Luo, Quan Sun, Yufeng Cui, Jinsheng Wang, Fan Zhang, Yueze Wang, Zhen Li, Qiying Yu, et~al.
\newblock Emu3: Next-token prediction is all you need.
\newblock {\em arXiv preprint arXiv:2409.18869}, 2024.

\bibitem{wang2025lavie}
Yaohui Wang, Xinyuan Chen, Xin Ma, Shangchen Zhou, Ziqi Huang, Yi~Wang, Ceyuan Yang, Yinan He, Jiashuo Yu, Peiqing Yang, et~al.
\newblock Lavie: High-quality video generation with cascaded latent diffusion models.
\newblock {\em International Journal of Computer Vision}, 133(5):3059--3078, 2025.

\bibitem{yu2023language}
Lijun Yu, Jos{\'e} Lezama, Nitesh~B Gundavarapu, Luca Versari, Kihyuk Sohn, David Minnen, Yong Cheng, Vighnesh Birodkar, Agrim Gupta, Xiuye Gu, et~al.
\newblock Language model beats diffusion--tokenizer is key to visual generation.
\newblock {\em arXiv preprint arXiv:2310.05737}, 2023.

\bibitem{zhang2023label}
Xiaoxuan Zhang and Krishna Garikipati.
\newblock Label-free learning of elliptic partial differential equation solvers with generalizability across boundary value problems.
\newblock {\em Computer Methods in Applied Mechanics and Engineering}, 417:116214, 2023.

\bibitem{zhu2018bayesian}
Yinhao Zhu and Nicholas Zabaras.
\newblock Bayesian deep convolutional encoder–decoder networks for surrogate modeling and uncertainty quantification.
\newblock {\em Journal of Computational Physics}, 2018.

\bibitem{zhuang2020comprehensive}
Fuzhen Zhuang, Zhiyuan Qi, Keyu Duan, Dongbo Xi, Yongchun Zhu, Hengshu Zhu, Hui Xiong, and Qing He.
\newblock A comprehensive survey on transfer learning.
\newblock {\em Proceedings of the IEEE}, 109(1):43--76, 2020.

\bibitem{zhuolumina}
Le~Zhuo, Ruoyi Du, Han Xiao, Yangguang Li, Dongyang Liu, Rongjie Huang, Wenze Liu, Xiangyang Zhu, Fu-Yun Wang, Zhanyu Ma, et~al.
\newblock Lumina-next: Making lumina-t2x stronger and faster with next-dit.
\newblock In {\em The Thirty-eighth Annual Conference on Neural Information Processing Systems}.

\end{thebibliography}

\clearpage

\appendix

\section{Limitations} \label{sec:appendix-limitation}

Despite the promising success of \name{}, we acknowledge that it has several key limitations. 

\textbf{Generalization}. Existing foundation models typically have zero-shot generalization capabilities. For example, CLIP \cite{radford2021learning}, which was pretrained on a large set of vision-language data, can perform zero-shot classification on images for domain-specific applications. While \name{} is trained on multiple datasets, generalizing to novel physical processes requires fine-tuning, as they may have unseen input channels or represent a drastically different dynamic system from those seen during training. We leave this to future work.

\textbf{Discretization Error}. The tokenization process introduces quantization errors, and while the refinement module helps mitigate this, residual errors can still affect the precision of long-term simulations. This is especially significant for datasets with low spatial or temporal variance which are much more sensitive to small perturbations. Exploring alternative tokenization schemes or end-to-end training of the tokenizer and autoregressive model could help minimize this error.

\textbf{Data Diversity}. \name{} was only trained on 2D datasets, due to the architecture of the video tokenizer. This limits its direct applicability to 3D physical systems or systems with significantly different spatial structures. Future work could explore more flexible tokenization architectures that enable the compression of higher spatial dimensions, and include data from outside The Well.

\section{Experimental settings} \label{sec:experimental-settings}
\textbf{Refinement Module }
For each trajectory in the raw training data, we randomly sample a starting timestamp and run autoregressive generation to obtain the training data for the refinement module. We adopted MSE loss. We use a global batch size of 64 frames, a learning rate of $5e-3$ and a cosine decay learning rate scheduler.  We trained each refinement model for 500 epochs on its respective data. Unlike the base model, which is trained in bfloat16 precision, we observe that using float32 precision is crucial to achieve high-quality outputs, especially for datasets with low spatial variance. 

\textbf{Tokenizer } We trained the universal tokenizer on the $8$ datasets in Table~\ref{tab:one_frame_prediction} for $1000$ epochs with an effective batch size of $32$. We optimize the models using AdamW~\citep{kingma2014adam} with a base learning rate of $1e-3$, using a $10$-epoch linear warmup, followed by a cosine decay schedule for the remaining $990$ epochs. For model selection, we average the validation loss across all datasets after each training epoch and use the model with the lowest validation loss as the final tokenizer checkpoint. 

\textbf{AR Model} For the autoregressive (AR) model, we trained for $10000$ steps with an effective batch size of $32$. We used Adam as the optimizer with a learning rate schedule similar to the tokenizer, where the number of warmup steps is set to $1000$. We validated the model after every $100$ training steps and used the best checkpoint for testing. For both tokenizer and AR training, we upsampled the smaller datasets to match the size of the largest one, ensuring the model learns from each dataset uniformly.

\textbf{Evaluation }
After training, we tested the model on the held-out test set provided by the Well~\citep{ohana2024well}. For the one-step setting, we evaluated the model on random sliding windows sampled from the test simulations. For the long-horizon setting, we always initiated the model from the beginning of each simulation. This adheres to the standard practice in the Well.

\textbf{Finetuning }
To adapt \name{} to an unseen task, we finetune both the tokenizer and the autoregressive model. Specifically, we finetune the tokenizer for $100$ epochs and the autoregressive model for $1000$ iterations, with similar learning rates and schedulers to pretraining. This means the compute requirement for each finetuning task is about $10\%$ of that of pretraining. Section~\ref{sec:adaptation} shows that \name{} was able to achieve strong performance even with this limited compute, demonstrating its usefulness for the broad research community.

\section{Compute resources} \label{sec:compute-resources}
We trained the tokenizer and \name{} on $8\times$ $40$GB A100 devices, and evaluated using $1\times$ $40$GB A100 device for each task. We trained \name{} for $24$ hours on $8\times$A100s for $8$ datasets. This is approximately equal to the combined cost of training the best baseline model for each dataset at current market rate cloud compute costs \footnote{Using pricing from Lambda Labs}. Each model in The Well required $12$ hours on $1\times$H100~\citep{ohana2024well}, for a total time of 96 H100 hours when only considering the best model for each dataset, or about half the A100 hours used by \name{}.


\section{Reproducibility statement} \label{sec:reproducibility}
We will release the training and evaluation code, as well as the model checkpoints. We also note that the Well's authors \footnote{https://github.com/PolymathicAI/the\_well/issues/49} reported some reproducibility issues with the baseline models at the moment and are planning to update the codebase and the paper. We cite the currently reported numbers in our main experiments. For numbers not reported (e.g. longer rollouts), we use the latest version of the official codebase at the time of writing.

\section{Licenses} \label{sec:licenses}
Cosmos \cite{agarwal2025cosmos} is licensed under Apache-2.0, and the Well \cite{ohana2024well} benchmark follows  BSD-3-Clause license. We respect the intended use of each artifact and complied with all license requirements.

\section{Statistical significance} \label{sec:stats}
While the Well does not publish variance of the baselines for test sampling, Table~\ref{tab:stat_sig_split_datasets_fullnames_ci} shows that our 95\% confidence interval for 1 frame prediction with \name{} is outside the range of the baseline mean assuming a normal distribution. For rollout predictions, we start from the beginning of each sequence and evaluate on the entire test dataset, just as the baseline was evaluated.
\begin{table}[h!]
\centering
\caption{\textbf{\name{} 1 frame prediction with 95\% confidence intervals}. }
\label{tab:stat_sig_split_datasets_fullnames_ci}
\resizebox{1.0\textwidth}{!}{%
\begin{tabular}{@{} p{5cm} >{\centering\arraybackslash}m{2cm}
                   p{5cm} >{\centering\arraybackslash}m{2cm} @{}}
\toprule
\textbf{Dataset} & \textbf{Interval} & \textbf{Dataset} & \textbf{Interval} \\
\midrule
\texttt{shear\_flow}                      & $0.070\pm 0.011$ & \texttt{turbulent\_radiative\_layer} & $0.210\pm .0344$\\
\addlinespace
\texttt{rayleigh\_benard}                 & $0.147\pm .029$& \texttt{gray\_scott\_reaction}  & $0.021\pm 0.005$ \\
\addlinespace
\texttt{acoustic\_scattering (maze)}      & $0.096\pm .002$ & \texttt{viscoelastic\_instability}        & $0.212\pm 0.029$ \\
\addlinespace
\texttt{active\_matter}                   & $0.090\pm 0.011$ & \texttt{helmholtz\_staircase}             & $0.018\pm 0.004$ \\
\bottomrule
\end{tabular}%
}
\end{table}

\section{Additional experiments}
\subsection{Pretrained vs scratch}
\begin{figure}[h]
    \centering
    \includegraphics[width=1.0\linewidth]{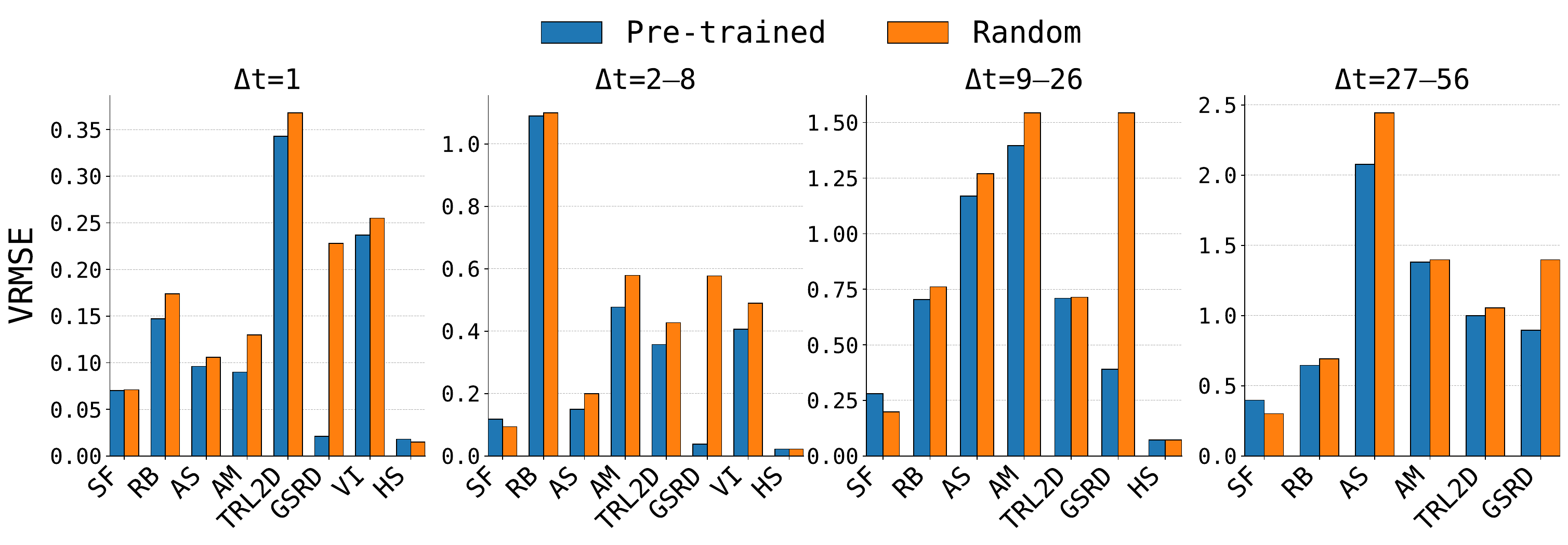}
    \caption{\textbf{Comparison of pretrained and randomly initialized weights}}
    \label{fig:initialization_comparison}
\end{figure}

Figure~\ref{fig:initialization_comparison} compares the performance of \name{} when initialized from a Cosmos pretrained checkpoint (Pre-trained) vs when initialized from scratch (Random). Using the pretrained checkpoint outperforms training from scratch across almost all tasks and evaluation settings, which shows the effectiveness of \name{} in transferring prior knowledge from natural videos to physical simulations. Table~\ref{tab:univ_vs_scratch} details the performance of the two models.

\begin{table}[h]
  \centering
  \small
  \caption{\textbf{Comparison of pre-trained and randomly initialized models.} Next-frame and long-horizon prediction results on the Well benchmark for Cosmos weights pre-trained on natural video and with randomly initialized weights.}
  \label{tab:univ_vs_scratch}
  \resizebox{1.0\textwidth}{!}{\begin{tabular}{@{}l cc cc cc cc@{}}
    \toprule
    \multirow{2}{*}{\texttt{Dataset}}
      &  \multicolumn{2}{c}{$\Delta t=1$}
      &  \multicolumn{2}{c}{$\Delta t=2\!:\!8$}
      &  \multicolumn{2}{c}{$\Delta t=9\!:\!26$}
      &  \multicolumn{2}{c}{$\Delta t=27\!:\!56$} \\
    \cmidrule(lr){2-3} \cmidrule(lr){4-5} \cmidrule(lr){6-7} \cmidrule(lr){8-9}
      & Pre.\ & Rand.
      & Pre.\ & Rand.
      & Pre.\ & Rand.
      & Pre.\ & Rand. \\
    \midrule
    \texttt{shear\_flow}
      & \textbf{0.070} & 0.071
      & 0.118          & \textbf{0.094}
      & 0.281          & \textbf{0.198}
      & 0.397          & \textbf{0.301} \\

    \texttt{rayleigh\_benard}
      & \textbf{0.147} & 0.174
      & \textbf{1.090} & 1.100
      & \textbf{0.704} & 0.761
      & \textbf{0.646} & 0.691 \\

    \texttt{acoustic\_scattering (maze)}
      & \textbf{0.096} & 0.106
      & \textbf{0.150} & 0.200
      & \textbf{1.170} & 1.270
      & \textbf{2.076} & 2.444 \\

    \texttt{active\_matter}
      & \textbf{0.090} & 0.130
      & \textbf{0.477} & 0.579
      & \textbf{1.396} & 1.544
      & \textbf{1.381} & 1.397 \\

    \texttt{turbulent\_radiative\_layer\_2D}
      & \textbf{0.343} & 0.368
      & \textbf{0.357} & 0.427
      & \textbf{0.710} & 0.714
      & \textbf{0.998} & 1.055 \\

    \texttt{gray\_scott\_reaction\_diffusion}
      & \textbf{0.021} & 0.228
      & \textbf{0.038} & 0.577
      & \textbf{0.390} & 1.544
      & \textbf{0.895} & 1.397 \\

    \texttt{viscoelastic\_instability}
      & \textbf{0.237} & 0.255
      & \textbf{0.406} & 0.490
      & —              & —
      & —              & — \\

    \texttt{helmholtz\_staircase}
      & 0.018          & \textbf{0.015}
      & 0.022          & 0.022
      & 0.072          & 0.072
      & —              & — \\
    \bottomrule
  \end{tabular}}
\end{table}


\subsection{Scaling results}
We study the scalability of \name{} by training and evaluating autoregressive models with $3$ different sizes: $700$M, $2$B, and $4$B. Since Cosmos only provides the $4$B model checkpoint, we initialized all $3$ models in this experiment from scratch for a fair comparison. Table~\ref{tab:scratch-performance-reorg} shows that $4$B is the best performing model, followed by $700$M, while $2$B performed the worst. We observed that both the $4$B and the $2$B models overfit whereas the $700$M model did not, and the $2$B model converged to a worse point compared to the $700$M and $4$B models, leading to overall poorer performances.

\begin{table}[h]
  \centering
  \small
    \caption{\textbf{Prediction errors for Scratch models at various time horizons.} We report next-frame and long-horizon prediction errors for Scratch 4B, Scratch 2B, and Scratch 700M across different datasets, highlighting the best (lowest) error in each horizon.}
  \label{tab:scratch-performance-reorg}
  \resizebox{1.0\textwidth}{!}{\begin{tabular}{@{}l ccc ccc ccc ccc@{}}
    \toprule
    \multirow{2}{*}{\texttt{Dataset}}
      & \multicolumn{3}{c}{$t+1$}
      & \multicolumn{3}{c}{$t+2\!:\!8$}
      & \multicolumn{3}{c}{$t+9\!:\!26$}
      & \multicolumn{3}{c}{$t+27\!:\!56$} \\
    \cmidrule(lr){2-4} \cmidrule(lr){5-7} \cmidrule(lr){8-10} \cmidrule(lr){11-13}
      & 4B & 2B & 700M
      & 4B & 2B & 700M
      & 4B & 2B & 700M
      & 4B & 2B & 700M \\
    \midrule
    \texttt{shear\_flow}
      & \textbf{0.071} & 0.075          & 0.073
      & \textbf{0.094} & 0.112          & 0.096
      & 0.198          & 0.216          & \textbf{0.166}
      & 0.301          & 0.303          & \textbf{0.257} \\

    \texttt{rayleigh\_benard}
      & \textbf{0.174} & 0.181          & 0.194
      & \textbf{1.10}  & 1.201          & 1.113
      & \textbf{0.761} & 0.855          & 0.827
      & \textbf{0.691} & 0.823          & 0.999 \\

    \texttt{acoustic\_scattering (maze)}
      & \textbf{0.106} & 0.110          & 0.120
      & \textbf{0.20}  & 0.211          & 0.237
      & 1.270          & 1.284          & \textbf{1.242}
      & 2.444          & 2.497          & \textbf{2.287} \\

    \texttt{turbulent\_radiative\_layer}
      & 0.368          & 0.421          & \textbf{0.312}
      & \textbf{0.427} & 0.443          & 0.450
      & \textbf{0.714} & 0.758          & 0.730
      & 1.055          & 1.099          & \textbf{0.942} \\

    \texttt{active\_matter}
      & 0.130          & \textbf{0.102} & 0.105
      & \textbf{0.579} & 0.592          & 0.623
      & 1.544          & 1.626          & \textbf{1.394}
      & \textbf{1.397} & 1.415          & 1.417 \\

    \texttt{gray\_scott\_reaction}
      & \textbf{0.228} & 0.230          & 0.231
      & 0.577          & \textbf{0.509} & 0.526
      & 1.544          & 1.126          & \textbf{1.051}
      & 1.397          & 2.290          & \textbf{1.300} \\

    \texttt{viscoelastic\_instability}
      & 0.255          & 0.319          & \textbf{0.246}
      & \textbf{0.490} & 0.494          & 0.590
      & —              & —              & —
      & —              & —              & — \\

    \texttt{helmholtz\_staircase}
      & 0.015          & 0.015          & \textbf{0.014}
      & 0.0224         & 0.019          & \textbf{0.017}
      & 0.0718         & \textbf{0.056} & 0.061
      & —              & —              & — \\
    \bottomrule
  \end{tabular}}
  \vspace{0.5em}
\end{table}

\subsection{More qualitative results}

\begin{figure}[t]
    \centering
    \includegraphics[width=1\linewidth]{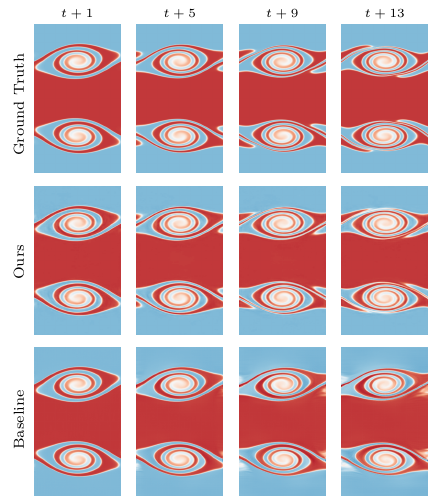}
    \caption{\textbf{Qualitative Comparisons on \texttt{shear\_flow} Dataset.} We compare the prediction of \name{}~ with the ground truth and the prediction of the best baseline model at lead times of 1,5,9,13 frames.}
    \label{fig:sf_qualitative_samples}
\end{figure}

\begin{figure}[t]
    \centering
    \includegraphics[width=1\linewidth]{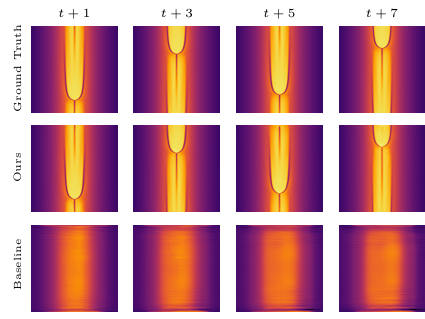}
    \caption{\textbf{Qualitative Comparisons on \texttt{viscoelastic\_instability} Dataset.} We compare the prediction of \name{}~ with the ground truth and the prediction of the best baseline model at lead times of 1,3,5,7 frames.}
    \label{fig:vi_qualitative_samples}
\end{figure}

\begin{figure}[t]
    \centering
    \includegraphics[width=1\linewidth]{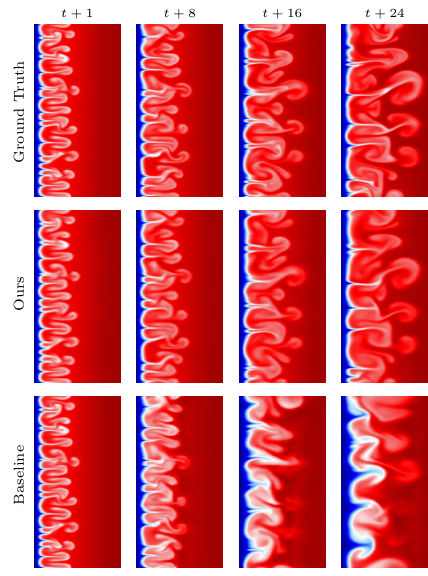}
    \caption{\textbf{Qualitative Comparisons on \texttt{rayleigh\_benard} Dataset.} We compare the prediction of \name{}~ with the ground truth and the prediction of the best baseline model at lead times of 1,8,16,24 frames.}
    \label{fig:rb_qualitative_samples}
\end{figure}

\begin{figure}[t]
    \centering
    \includegraphics[width=1\linewidth]{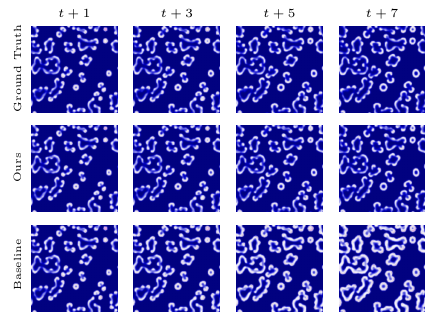}
    \caption{\textbf{Qualitative Comparisons on \texttt{gray\_scott\_reaction\_diffusion}  Dataset.} We compare the prediction of \name{}~ with the ground truth and the prediction of the best baseline model at lead times of 1,3,5,7 frames.}
    \label{fig:gs_qualitative_samples}
\end{figure}

We provide additional visualizations of the \name{}'s prediction results on \texttt{shear\_flow} (Figure \ref{fig:sf_qualitative_samples}), \texttt{viscoelastic\_instability} (Figure \ref{fig:vi_qualitative_samples}), \texttt{rayleigh\_benard} (Figure \ref{fig:rb_qualitative_samples}) and \texttt{gray\_scott\_reaction\_diffusion} (Figure \ref{fig:gs_qualitative_samples}). We compare the prediction of \name{}~with the ground truth and the prediction of baseline models at various lead times. \name{}~shows consistent improvement over baselines across all lead times. The improvements on longer lead times are more pronounced.

\end{document}